\pdfoutput=1
\def\year{2019}\relax
\documentclass[letterpaper]{article} 
\usepackage{aaai19}  
\usepackage{times}  
\usepackage{helvet}  
\usepackage{courier}  
\usepackage{url}  
\usepackage{graphicx}  

\usepackage{amsmath}
\usepackage{amsfonts}
\usepackage{booktabs}
\usepackage[linesnumbered,ruled]{algorithm2e}
\usepackage{adjustbox}
\usepackage{subfigure}
\usepackage{multicol}

\newcommand{\DPatch}{\textsc{DPatch}}
\begin{document}
%
\title{\DPatch: An Adversarial Patch Attack on Object Detectors}

\author{Xin Liu$^1$, Huanrui Yang$^1$, Ziwei Liu$^2$, Linghao Song$^1$, Hai Li$^1$, Yiran Chen$^1$\\
    \large $^1$Duke University,~
    $^2$The Chinese University of Hong Kong\\
    $^1$\{xin.liu4, huanrui.yang, linghao.song, hai.li, yiran.chen\}@duke.edu,~
    $^2$zwliu.hust@gmail.com
}

\maketitle
\begin{abstract}
Object detectors have emerged as an indispensable module in modern computer vision systems.
In this work, we propose \DPatch -- a black-box adversarial-patch-based attack towards mainstream object detectors (\textit{i.e.} Faster R-CNN and YOLO).
Unlike the original adversarial patch that only manipulates image-level classifier, our \DPatch\ simultaneously attacks the bounding box regression and object classification so as to disable their predictions.
Compared to prior works, \DPatch\ has several appealing properties:
(1) \DPatch\ can perform both untargeted and targeted effective attacks, degrading the mAP of Faster R-CNN and YOLO from 75.10\% and 65.7\% down to below 1\%, respectively;
(2) \DPatch\ is small in size and its attacking effect is location-independent, making it very practical to implement real-world attacks;
(3) \DPatch\ demonstrates great transferability among different detectors as well as training datasets. For example, \DPatch\ that is trained on Faster R-CNN can effectively attack YOLO, and vice versa. 
Extensive evaluations imply that \DPatch\ can perform effective attacks under black-box setup, i.e., even without the knowledge of the attacked network's architectures and parameters.
Successful realization of \DPatch\ also illustrates the intrinsic vulnerability of the modern detector architectures to such patch-based adversarial attacks.
\end{abstract}

\section{Introduction}
As deep learning systems achieve excellent performance in many cognitive applications, their security and robustness issues are also raised as important concerns recently. 
Among them, object detector is an indispensable module in modern computer vision systems, and widely deployed in surveillance~\cite{liu2015deep} systems and autonomous vehicles~\cite{FastRCNN}. 
It becomes an increasing vital task to comprehensively study their vulnerability to adversarial attacks~\cite{nicolas2017practical,fangzhou2017defense}.

Many existing adversarial attacks focus on learning full-image additive noise, where the predictions of deep learning system are manipulated by injecting small perturbations into the input samples.
Though the injected noises are small and sometimes even invisible to human eyes, these methods need to manipulate the whole image and become less impractical for the real-world physical attacks. 

Recently, adversarial patch~\cite{AdvPatch} is introduced as an practical approach of real-world attacks. 
The adversarial patch misleads a CNN classifier to predict any object that coupled with the pasted patch to a targeted class, regardless of the object's scale, position, and direction.
The effectiveness of such adversarial attacks has been also proven in a black-box setup where the structure and parameters of the attacked neural networks are unknown to the attacker.

\begin{figure}[t]
\centering
\includegraphics[width=0.95\columnwidth]{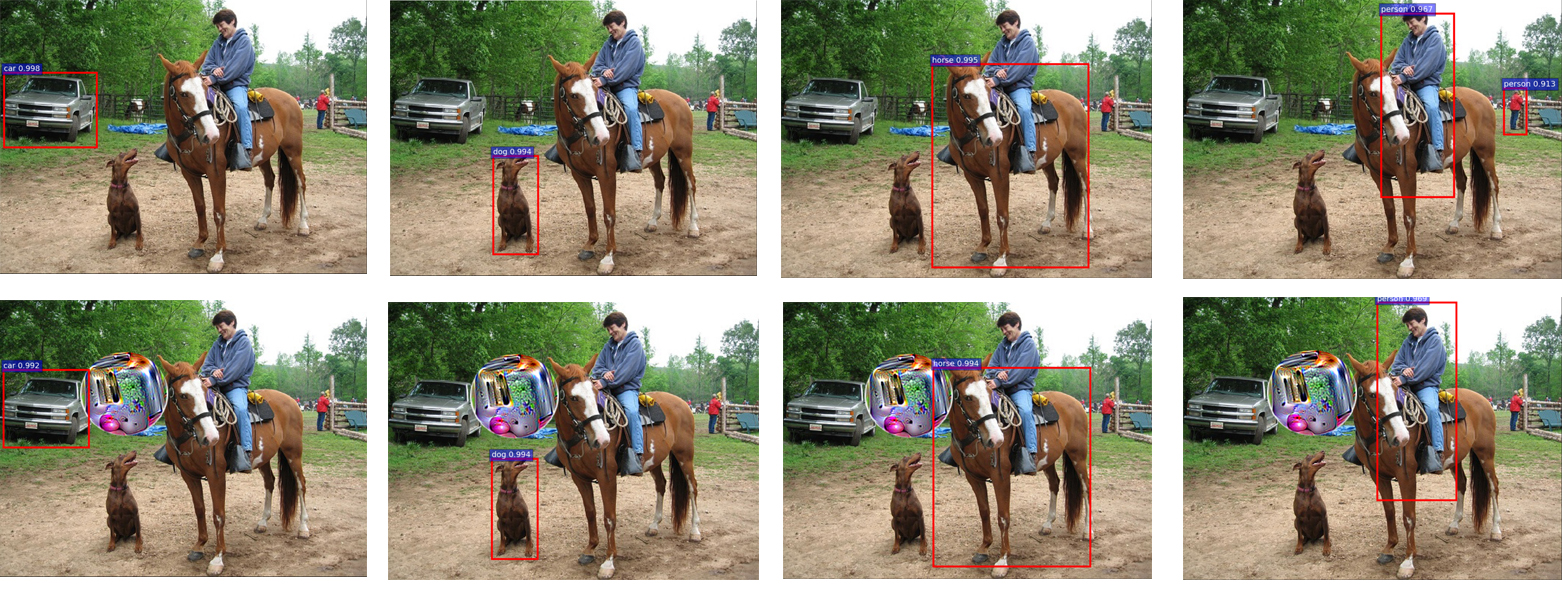}
\vspace{-10pt}
\caption{The original adversarial patch fails to attack object detectors. The first row is the original image. Faster R-CNN can detect multiple objects in the scene with a high accuracy. The second row is the image embedded with the Google's adversarial patch whose targeted class is $toaster$. Faster R-CNN is not influenced by the patch and can still be able to correctly recognize all the objects. 
}
\label{fig1_compare}
\vspace{-10pt}
\end{figure}

However, we found that the original adversarial patch technique is not able to fool object detectors such as Faster R-CNN \cite{FasterRCNN} and YOLO \cite{redmon2016yolo9000}, as shown in Fig.\ref{fig1_compare}.
The reason resides at the detector architectures:
Modern object detectors first locate the objects with different sizes at different locations on the image and then perform classification. Hence, the number of targets that need to be attacked in this case is much larger than that in a pure classification application~\cite{AdvFace}. 
For example, Faster R-CNN generates approximately 20k regions for classification, which is far beyond an original adversarial patch can effectively attack.

Our key insight is that both bounding box regression and object classification need to be simultaneously attacked.
Based on this observation, we propose \DPatch\ -- an iteratively trained adversarial patch that effectively attacks mainstream object detectors. A small \DPatch, e.g., a 40-by-40 pixel region, can perform both untargeted attack and targeted attack:
In an untargeted attack, the object detector cannot locate the correct region containing normal object; 
in a targeted attack, the object detector can only detect the \DPatch\ but ignore any other objects on the image. 

Extensive evaluations imply that \DPatch\ can perform effective attacks under black-box setup, i.e., even without the knowledge of the attacked network's architectures and parameters.
Successful realization of \DPatch\ also illustrates the intrinsic vulnerability of the modern detector architectures to such patch-based adversarial attacks.

\section{Related Work}

\noindent\textbf{Attacking Deep Learning Systems.}
Convolutional neural networks (CNNs) have been proven vulnerable to so called adversarial attack, where the classification result of the CNN can be manipulated by adding small perturbations onto its input examples~\cite{Intriguing,kevin2018robust,chaowei2018spatially}.
The adversarial examples can be crafted by gradient based algorithms such as Fast Gradient Sign Method (FGSM) \cite{Explaining} and Projected Gradient Descent (PGD) \cite{Towardsdeep} or iterative methods such as DeepFool \cite{Deepfool} and Carlini-Wagner (CW) attack \cite{Toevaluating}. 
Based on the concept of adversarial attack, an ``adversarial glasses'' \cite{Accessorize} was designed to fool a face recognition system so that the system recognizes the wearer as someone else. 
Most of the previous stuides on adversarial attack are about pixel-wise additive noise \cite{Explaining,Deepfool,Toevaluating,chaowei2018spatially}. 
However, these attacks change all the pixels of the input image by a small amount, which tends to be not feasible in real-world vision systems such as web cameras and autonomous vehicles.

\noindent\textbf{Adversarial Patch.}
To achieve a universal attack on real-world vision system, 
Google \cite{AdvPatch} recently designed a universal, robust adversarial patch that can be applied in real physical scene, causing a classifier to output any targeted class \cite{AdvPatch}. 
Based on the principle that object detection networks are used to detecting the most "salient" object in an image, the adversarial patch can be trained to be more "salient" than other objects in the scene. Specifically, the patch is  trained to optimize the expected probability of a target class \cite{AdvPatch}. 
During the training process, the patch is applied to the scene at a random position with a random scale and rotation. This process enables the patch to be robust against shifting, scaling and rotating. Finally, the object in the scene will be detected as a targeted class of the patch.

\begin{figure*}[htb]
\centering
\includegraphics[width=1.8\columnwidth]{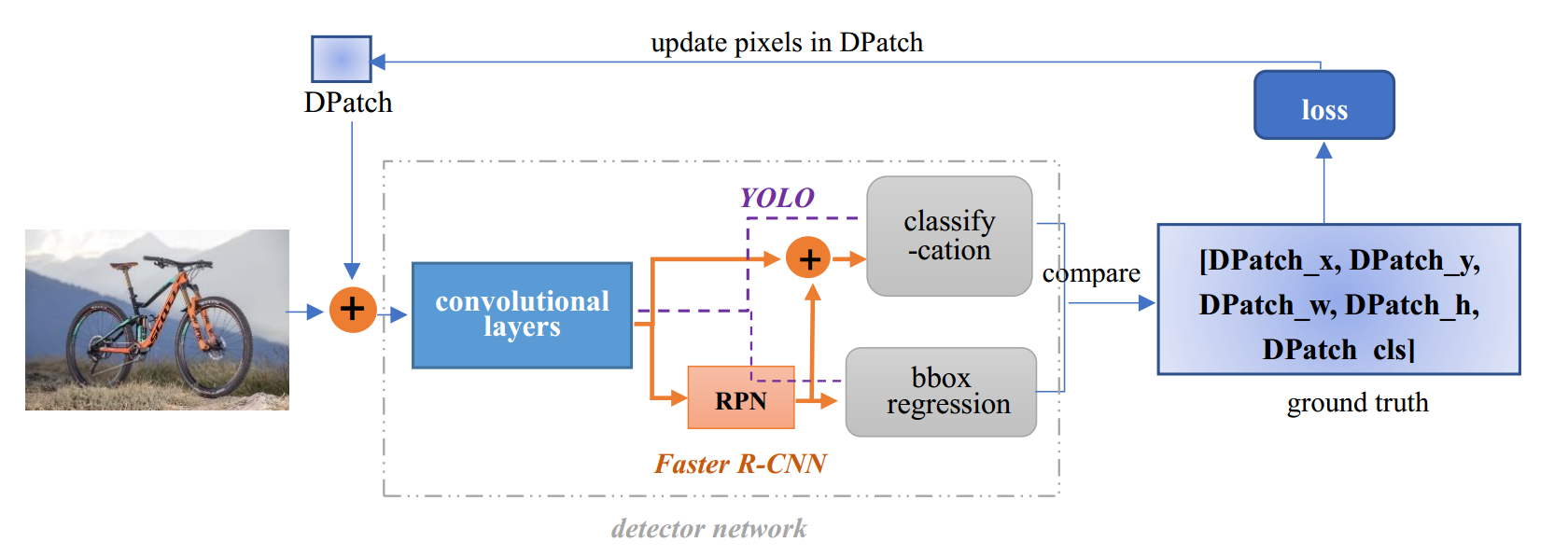}
\vspace{-15pt}
\caption{\DPatch\ training system: we add a randomly-initialized \DPatch\ to the image, utilize the detector network to do classification and bounding box regression based on the ground truth \textbf{[\DPatch\_x, \DPatch\_y, \DPatch\_w, \DPatch\_h, target\_label]}. During back-propagation, we update the pixels of \DPatch. }
\label{fig2_training_system}
\vspace{-15pt}
\end{figure*}

\section{Proposed Approach}
\subsection{Revisit Modern Detectors}

\noindent\textbf{Faster R-CNN.}
As mentioned before, Faster R-CNN is a typical two-stage detector. The first stage is to propose regions through a deep fully convolutional network, while the second stage is the Fast R-CNN detector that uses the proposed regions.

\begin{description}
  \item[$\bullet$  Region Proposal Network] After extracting features by convolutional layers, Faster R-CNN uses Region Proposal Network(RPN) to generate the region proposals that the object probably lies in. This Region Proposal Network takes as input an $n\times n$ spatial window ($3\times 3$ window as default) to slide over the last shared convolutional layer. These proposed regions generated by RPN will be mapped to the previous feature map for object classification later on.
   
    \item[$\bullet$  Anchors Generation and Region of Interest] 
    Faster R-CNN uses a $3\times 3$ window to slide over the last shared convolutional feature map. An anchor is centered at the sliding window. It generates 9 anchor boxes at each sliding position associated with 3 scales and 3 aspect ratios. These anchor boxes are considered as the potential proposal. When evaluating, Faster R-CNN uses a \textit{cls} layer, which outputs \textit{2k} scores to estimate probability of object or not for each potential proposal. If the cls score is comparatively high, the potential proposal will be considered as the Region of Interest(RoI) where an object exist. Otherwise the proposal will be ignored.
    
    In order to attack the 2-stage detector like Faster R-CNN, it is necessary to invalidate the generation of correct region proposal. Our purpose is to make the region where the \DPatch\ exists as the only valid RoI, while other potential proposal should be considered not to own an object and thus, ignored. 

\end{description}

\noindent\textbf{YOLO.}
YOLO is introduced as a unified model for object detection. It reframes object detection as a single regression problem, straight from image pixels to bounding box coordinates and class probabilities. YOLO has great advantages of fast detection and high mAP.

\begin{description}
  \item[$\bullet$ Unified Detection] Unlike Faster R-CNN, YOLO is a single convolutional network simultaneously predicts multiple bounding boxes and class probabilities for those boxes. Specifically, it divides the input image into multiple grids, predicts bounding boxes and confidence scores for each grid. These confident scores reflect the confidence of the box containing an object, as well as the accuracy of the model predicting that box. This one-stage detection system trains on full images and directly optimizes detection performances \cite{yolo}. By such optimized design, YOLO can run at 45 frames per second on a Titan X GPU and also achieve more than twice the mean average precision (mAP) of other real-time detection systems.

  \item[$\bullet$ Bounding Boxes Prediction and Confidence Scores] Each grid in the image predicts \textit{B} bounding boxes and  confidence scores for those boxes. These confidence scores reflect how confident that the box contains an object and how accurate the box is. If the confidence score is comparatively lower, that bounding box predicted by the grid will not be considered to contain a real object. 
  Similarly, the grids where the \DPatch\ exists should be considered to have an object when attacking YOLO, while other grids should be ignored. That is, the grid containing a \DPatch\ has higher confidence score than others with normal objects.
\end{description}

\subsection{\DPatch\ Formulation} 

\noindent\textbf{Original Adversarial Patch.} 
As introduced in the Introduction section, the adversarial patch proposed by Google~\cite{AdvPatch} is aiming for maximizing the loss of a CNN classifier when the patch is applied to an input image. To make the patch effective on all the inputs and under potential transformations in the physical world, the patch is optimized over the expectation over random input images, locations and transformations. 
\vspace{-5pt}
\begin{equation}
\label{equ:advpatch}
    \hat{P} = arg\max_P \mathbb{E}_{x,t,l}[log Pr(\hat{y}|A(P,x,l,t))]
    \vspace{-5pt}
\end{equation}
Equation~(\ref{equ:advpatch}) gives a formal description of the training objective of adversarial patch~\cite{AdvPatch}, where $A(P,x,l,t)$ denotes the input image produced by applying patch $P$ to the original image $x$ at location $l$ with transformation $t$. Potential transformation includes scaling and rotating the patch. $Pr(\hat{y}|A)$ is the probability of classifying input $A$ into the true label $\hat{y}$ providing the CNN classifier. Optimizing this objective could produce a patch that can effectively attack the classifier invariant of shifting, scaling and rotating when applied to any input. \\

\noindent\textbf{Adversarial Patch on Object Detectors.}
Inspired by the Google adversarial patch, we propose \DPatch, an adversarial patch that works against state-of-the-art object detectors. We design \DPatch\ for both untargeted and targeted attack scenarios. For training the untargeted \DPatch, we would like to find a patch pattern $\hat{P_u}$ that maximize the loss of the object detector to the \emph{true} class label $\hat{y}$ and bounding box label $\hat{B}$ when the patch is applied to the input scene $x$ using ``apply'' function $A$, as shown in equation~(\ref{equ:untarget}). And in a targeted attack setting, we would like to find a patch pattern $\hat{P_t}$ that minimize the loss to the \emph{targeted} class label $y_t$ and bounding box label $B_t$, as shown in equation~(\ref{equ:target}).
\vspace{-5pt}
\begin{equation}
\label{equ:untarget}
    \hat{P_u} = arg\max_P \mathbb{E}_{x,s}[L(A(x,s,P);\hat{y},\hat{B})]
\end{equation}
\vspace{-10pt}
\begin{equation}
\label{equ:target}
    \hat{P_t} = arg\min_P \mathbb{E}_{x,s}[L(A(x,s,P);y_t,B_t)]
    \vspace{-5pt}
\end{equation}

The "apply" function $A(x,s,P)$ means adding patch $P$ onto input scene $x$ with shift $s$. Here we uniformly sample the shift $s$ within the scene during the training to make our patch shift invariant.

\begin{figure} 
  \centering 
  \subfigure[a 40-by-40 untargeted \DPatch\ that aims to attack YOLO]{ 
    \label{fig_patch_a} 
    \includegraphics[width=1.4in]{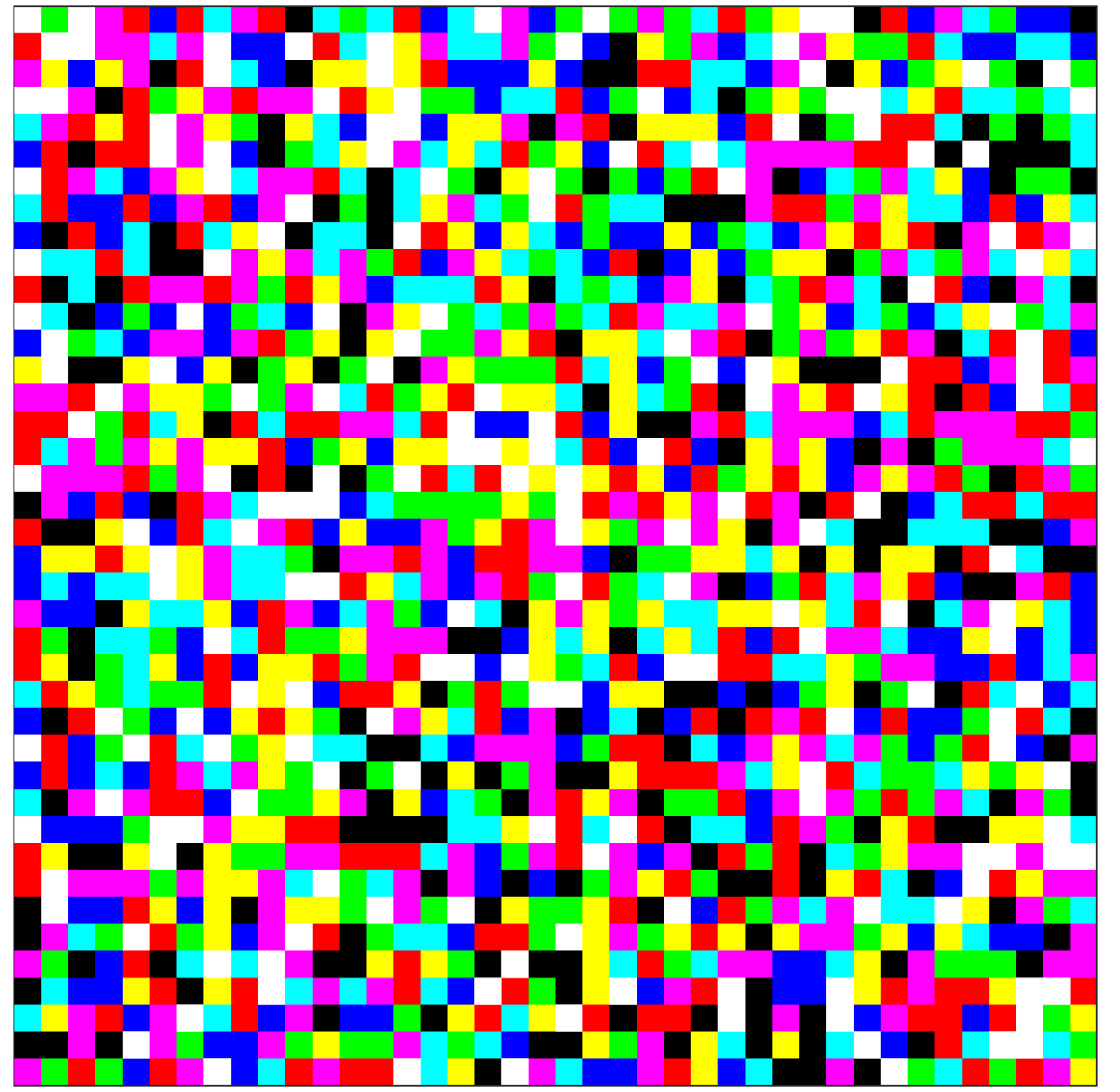}} 
  \hspace{0.1in} 
  \subfigure[A 20-by-20 $tv$ targeted \DPatch\ that attacks Faster R-CNN]{ 
    \label{fig_patch_b} 
    \includegraphics[width=1.4in]{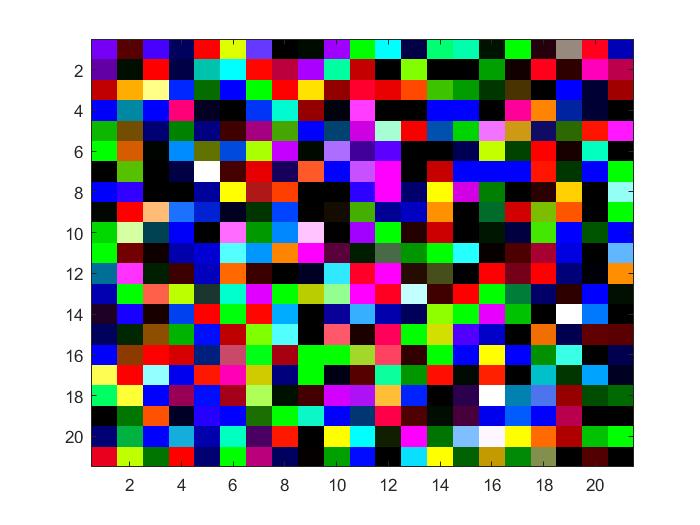}} 
\vspace{-5pt}
  \caption{two \DPatch\ es. } 
  \label{fig_patch} 
\vspace{-15pt}
\end{figure}

\subsection{Object Detector Attacking System}

\noindent\textbf{\DPatch\ Training.}
As shown in Fig.\ref{fig2_training_system}, we add a randomly initialized \DPatch\ pattern into the input image before it enters the detector system, and define the ground truth as [\DPatch\_x, \DPatch\_y, \DPatch\_w, \DPatch\_h, target\_label]. \DPatch\_x and \DPatch\_y represent the location of the \DPatch\ while \DPatch\_w and \DPatch\_h represent its width and height. The bounding box label is [\DPatch\_x, \DPatch\_y, \DPatch\_w, \DPatch\_h], similar to the ground truth of detector's bounding boxes. For untargeted attack, target\_label is defined as 0; for targeted attack, target\_label is equal to the label of targeted class. Note that the detector system can either be Faster R-CNN or YOLO. 
The basic \DPatch\ training process starts with a pretrained Faster R-CNN or YOLO using Pascal VOC 2007 dataset (see details in \textbf{Experiments}). \\

\noindent\textbf{\DPatch\ Design.}
Our default \DPatch\ is a 40-by-40 square which is attached onto the top left corner of the image. 
\begin{description}
  \item[$\bullet$ Randomly-located \DPatch]
In order to analyze the influences of different locations and to make \DPatch\ shift invariant, we randomly shift the location of the \DPatch\ while keeping all the pixels in the \DPatch\ same. Specifically, we randomly initialize the value of shift \textit{s} when preparing \DPatch\ at each iteration of training, but leaving the pixel points in it unmodified. In such a case, each image in the dataset will be attached with the same \DPatch\ but at different locations. The size of randomly-located \DPatch\ is still 40-by-40, the same as the default setting.

\item[$\bullet$ \DPatch\ of Different Targeted Classes]
As there exist more than 10 object classes in the dataset used to train a detector, e.g., 20 classes in Pascal VOC 2007, it is intriguing to see whether mAP will fall down to a similar value if we set the \DPatch\ with different labels. For example, for targeted attack, we can set the ground truth as [0, 0, 40, 40, 15], 15 is the label index representing \textit{person}. That is, we hope that the detector can only recognize the \DPatch\ and classified the \DPatch\ as \textit{person}. In our experiment, we randomly select four classes, \textit{bike}, \textit{boat}, \textit{cow} and \textit{tv} from Pascal VOC 2007 and evaluate their attack effects.

\item[$\bullet$ \DPatch\ with Different Sizes]
\DPatch\ size is another significant predetermined factor that could affect the effectiveness of \DPatch\ attack. There is a tradeoff between smaller patches that are harder to detect and defense, while larger patches that provide better attacking effect. In our experiment, we produce three different sizes of \DPatch\ es, namely, 20-by-20, 40-by-40 and 80-by-80 to test the efficiency of their attacks. In this way, we can better understand the relationship between \DPatch\ sizes and their attacking effects, so that we can find the minimal possible size of a patch for meaningful attack.
\end{description}

\subsection{Transferability of \DPatch}
Usually it is unknown that a detector is trained by which architecture. In such case, it makes more sense that the \DPatch\ trained by YOLO can also fool Faster R-CNN, or the \DPatch\ trained by Faster R-CNN with different classifiers can fool each other. To verify this idea, we train a \DPatch\ via YOLO and then add this trained \DPatch\ to input image, let Faster R-CNN work as the detector. If Faster R-CNN cannot correctly detect the object in the input image, our \DPatch\ attack should be considered as transferrable among detectors. Moreover, we train a \DPatch on COCO and then attack a Pascal VOC trained detector. If the mAP obviously decreased, \DPatch should be considered as transferrable between datasets.

\begin{figure}[t] 
  \centering 
  \subfigure[No \DPatch]{ 
    \label{fig:subfig:a} 
    \includegraphics[width=1.5in]{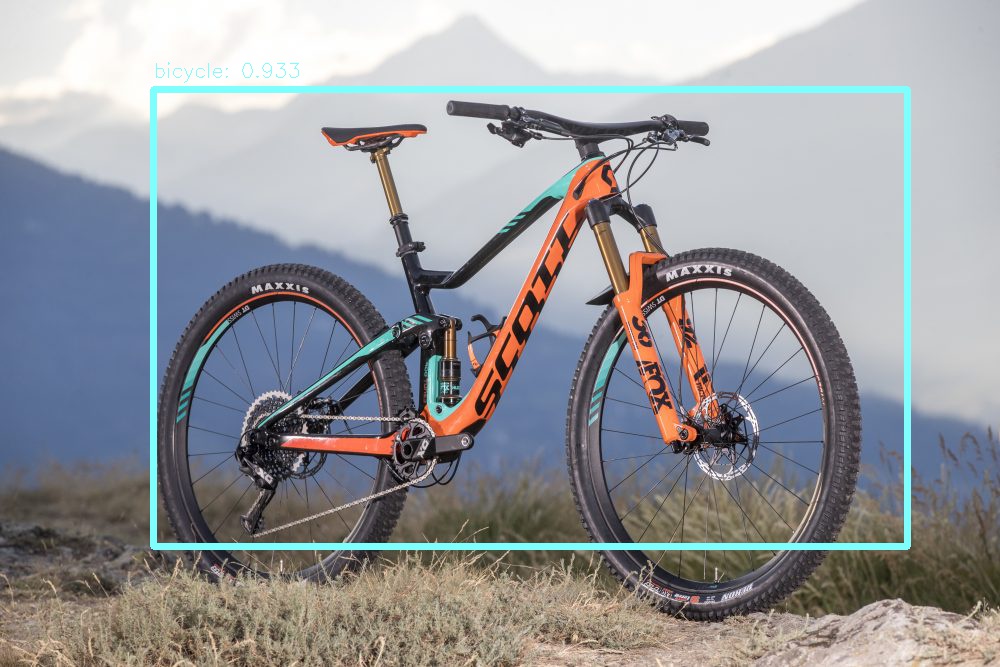}} 
  \hspace{0.05in} 
  \subfigure[With \DPatch]{ 
    \label{fig:subfig:b} 
    \includegraphics[width=1.5in]{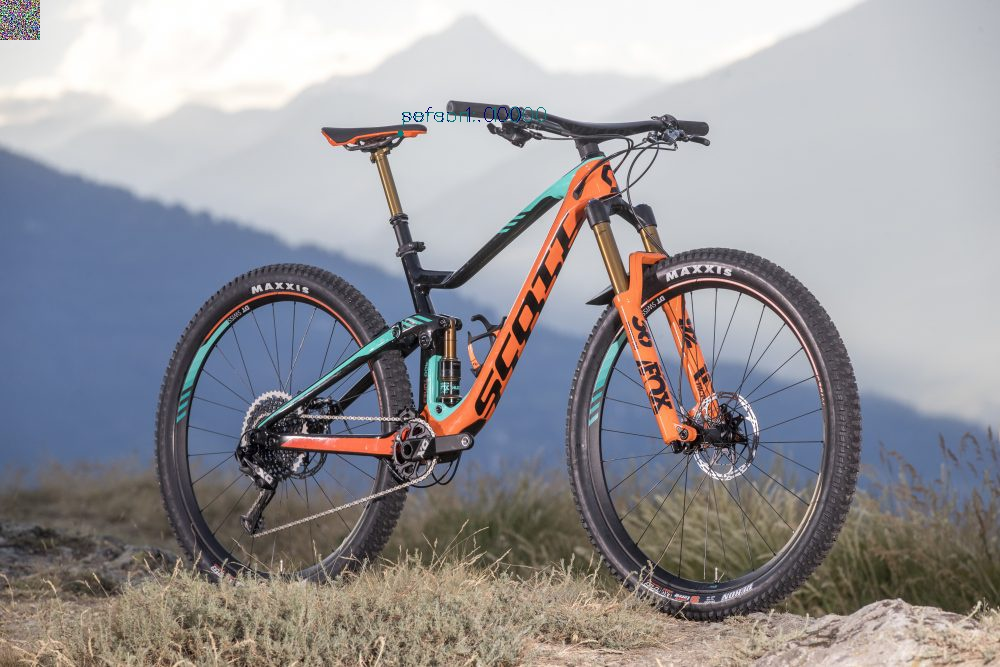}} 
\vspace{-10pt}
  \caption{YOLO cannot detect $bike$ after adding \DPatch\  Either the predicted bounding box or the classification result is incorrect. The predicted bounding box is just a dot, while the classification result is $sofa$ and $person$. } 
  \label{fig6} 
\vspace{-10pt}
\end{figure}

\section{Experiments}

\noindent\textbf{Setup.}
We use the pretrained YOLO and Faster R-CNN with VGG16 and ResNet101 as basic networks. Pascal VOC 2007\cite{pascal-voc-2007} is utilized as the dataset, which includes 20 labelled object classes. For YOLO, the average mAP of the 20 classes is 65.7\%; for Faster R-CNN with VGG16, the average mAP is 70.01\%, for Faster R-CNN with ResNet101, the average mAP is 75.1\%. Most previous work get similar detection mAP values as ours, so we can use these three networks as our pretrained detector networks. When performing targeted attack, we randomly choose one class from Pascal VOC 2007 to be our target class. Note that the default setting of YOLO is to generate $7\times 7$ grids in each image, however, our experiment generate $10\times 10$ grids instead. This change does not affect the overall detection performance of YOLO.

\subsection{Untargeted \DPatch\ Attack}

Fig.\ref{fig_patch_a} shows a 40-by-40 \DPatch\ that is designed as an untargeted adversarial patch. This \DPatch\ just aims to disable the detector(YOLO here), while it does not have a specific label. That is, it is unnecessary for the detector to accurately locate this \DPatch\ and to classify it as some class. As demonstrated, the original image Fig.\ref{fig_patch_b} where the bike should be correctly detected by YOLO, is detected wrongly as sofa and person in Fig.\ref{fig6}(b). 

After adding a untargeted \DPatch\ to each image in the testing dataset, mAPs of all classes are decreased obviously. As Table.\ref{table:faster_rcnn} and Table. \ref{table:yolo} show, For YOLO, mAP is decreased from 65.7\% to 0. For Faster R-CNN with ResNet101, mAP is decreased from 75.10\% to 0.

\begin{table*}[htb!]
\caption{Results on Pascal VOC 2007 test set with Fast R-CNN and ResNet101 when applying \DPatch\ of different types}
\small
\centering
\vspace{0.1cm}
\begin{adjustbox}{width=1\textwidth}
\begin{tabular}{p{4.25cm}p{1.2cm}p{1.2cm}p{1.2cm}p{1.2cm}p{1.2cm}p{1.2cm}p{1.2cm}p{1.2cm}p{1.2cm}p{1.2cm}p{1.2cm}} 
\hline \hline
\bf{Faster R-CNN}	&plane 	&bike 	&bird 	&boat 	&bottle	&bus	&car 	&cat	&chair 	&cow 	&table
\\
no \DPatch	&74.80 	&80.20 	&77.60 	&64.50 	&61.50 	&81.10 	&86.70 	&86.40 	&55.70 	&89.30 	&69.60 
\\
untargeted \DPatch	&0.10 	&3.20 	&4.30 	&0.00 	&5.40 	&0.00 	&9.80 	&0.00 	&11.20 	&10.60 	&5.20 
\\
targeted \DPatch	&0.02 	&0.00 	&0.00 	&0.00 	&0.00 	&0.53 	&0.08 	&0.61 	&0.00 	&0.02 	&0.00 
\\
YOLO trained \DPatch	&2.27 	&\textbf{0.51} 	&\textbf{0.87} 	&2.27 	&\textbf{0.78} 	&1.52 	&4.55 	&0.62 	&1.17 	&3.03 	&2.10 
\\ \hline
    &dog	&horse	&motor	&person	&plant	&sheep	&sofa	&train	&tv	& \multicolumn{2}{l}{mAP}
\\
&87.40 	&84.50 	&80.00 	&78.60 	&47.70 	&76.00 	&74.60 	&76.60 	&73.70 	&75.10 
\\
&0.30 	&0.59 	&0.00 	&1.69 	&0.00 	&4.68 	&0.00 	&0.00 	&1.00 	&\textbf{2.90} 
\\
&9.09 	&0.16 	&0.00 	&9.09 	&0.16 	&0.00 	&9.09 	&0.00 	&0.00 	&\textbf{0.98} 
\\
&2.02 	&3.37 	&1.30 	&0.94 	&0.53 	&0.43 	&3.03 	&1.52 	&1.52 	&\textbf{1.72} 
\\ \hline

\end{tabular}
\end{adjustbox}
\setlength{\belowcaptionskip}{5pt}
\label{table:faster_rcnn}
\vspace{-0.3cm}
\end{table*}

\begin{table*}[htb!]
\caption{Results on Pascal VOC 2007 test set with YOLO when applying \DPatch\ of different types }
\small
\centering
\vspace{0.1cm}
\begin{adjustbox}{width=1\textwidth}
\begin{tabular}{p{4.25cm}p{1.2cm}p{1.2cm}p{1.2cm}p{1.2cm}p{1.2cm}p{1.2cm}p{1.2cm}p{1.2cm}p{1.2cm}p{1.2cm}p{1.2cm}} 
\hline \hline
\bf{YOLO}    &plane 	&bike 	&bird 	&boat 	&bottle	&bus	&car 	&cat	&chair 	&cow 	&table     
\\ 
no \DPatch\	&69.50 	&75.60 	&64.00 	&52.30 	&35.60 	&73.40 	&74.00 	&79.60 	&42.10 	&66.10 	&66.90 
\\
untargeted \DPatch\	&0.00 	&1.50 	&9.10 	&1.30 	&9.10 	&0.00 	&9.10 	&0.00 	&9.10 	&9.10 	&0.40 
\\
targeted \DPatch\	&0.00 	&4.55 	&9.09 	&0.00 	&0.09 	&0.00 	&9.09 	&1.82 	&0.01 	&0.00 	&0.36 
\\
Faster R-CNN trained \DPatch\	&0.01 	&\textbf{0.00} 	&\textbf{0.23} 	&0.02 	&\textbf{0.00} 	&0.00 	&0.00 	&0.00 	&0.01 	&0.00 	&0.00 
\\  \hline

    &dog	&horse	&motor	&person	&plant	&sheep	&sofa	&train	&tv	&\multicolumn{2}{l}{mAP}
\\ 

&78.10 	&80.10 	&78.20 	&65.90 	&41.70 	&62.00 	&67.60 	&77.60 	&63.10 	&65.70 
\\
&0.00 	&0.00 	&0.00 	&0.00 	&9.10 	&9.10 	&0.00 	&0.00 	&1.00 	&\textbf{0.00} 
\\
&0.01 	&0.00 	&0.00 	&1.73 	&0.00 	&0.00 	&1.07 	&0.00 	&9.09 	&\textbf{1.85} 
\\
&0.00 	&0.03 	&0.00 	&0.07 	&0.00 	&0.00 	&0.00 	&0.00 	&0.01 	&\textbf{0.02} 
\\ \hline

\end{tabular}
\end{adjustbox}
\setlength{\belowcaptionskip}{5pt}
\label{table:yolo}
\vspace{-0.3cm}

\end{table*}

\subsection{Targeted \DPatch\ Attack}

\begin{figure}[t]
\centering
\includegraphics[width=0.95\columnwidth]{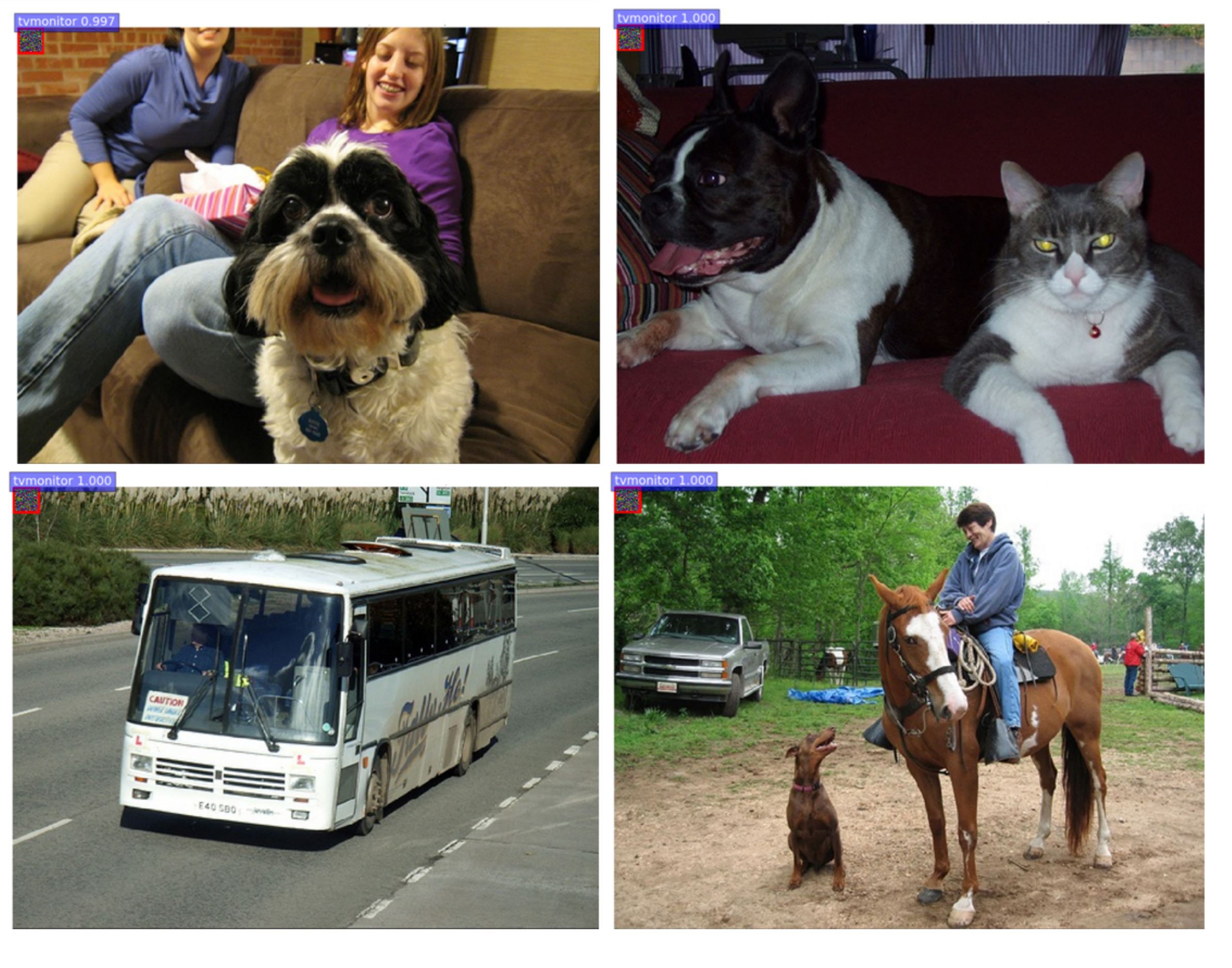}
\vspace{-10pt}
\caption{The \DPatch\ is placed on the \textbf{left top corner} of the images. Faster R-CNN networks can on longer recognize other objects. Here, we apply a  \DPatch\ whose targeted class is $tv$, it can fully fool the Faster R-CNN networks though the \DPatch\ size is comparatively small compared to other objects in the scene.}
\label{fig_targeted_fasterrcnn}
\vspace{-5pt}
\end{figure}

\noindent\textbf{Fixed-sized and Fixed-located \DPatch.}
Fig.\ref{fig_patch_b} shows a 20-by-20 \DPatch, designed to attack Faster R-CNN with ResNet101, whose targeted class is $tv$. This $tv$ \DPatch\ aims to make its occupied region to be recognized as the only valid RoI by Faster R-CNN. In 
Fig.\ref{fig_targeted_fasterrcnn}, our \DPatch\ covers the left top corner of each image, though the patch size(20-by-20) is small compared to other objects in the scene, it successfully fools the Faster R-CNN classifier and make it yield one result: the targeted class of the \DPatch. Hence, the function of multi-object detection and recognition of Faster R-CNN models is invalidated. The predicted probability is 0.997 of the first image of Fig.\ref{fig_targeted_fasterrcnn}, the predicted probability is 1.000 of the other three images. These predictions make sense because only the \DPatch\ occupied region can be recognized, other regions are ignored.

Specifically, the main purpose of applying such \DPatch\ is to make mean Accuracy Precision (mAP) of all classes in the dataset drop down to a lower value. The more mAP decreases, more successful the \DPatch\ attack is.

\begin{figure}[t]
\centering
\includegraphics[width=0.80\columnwidth]{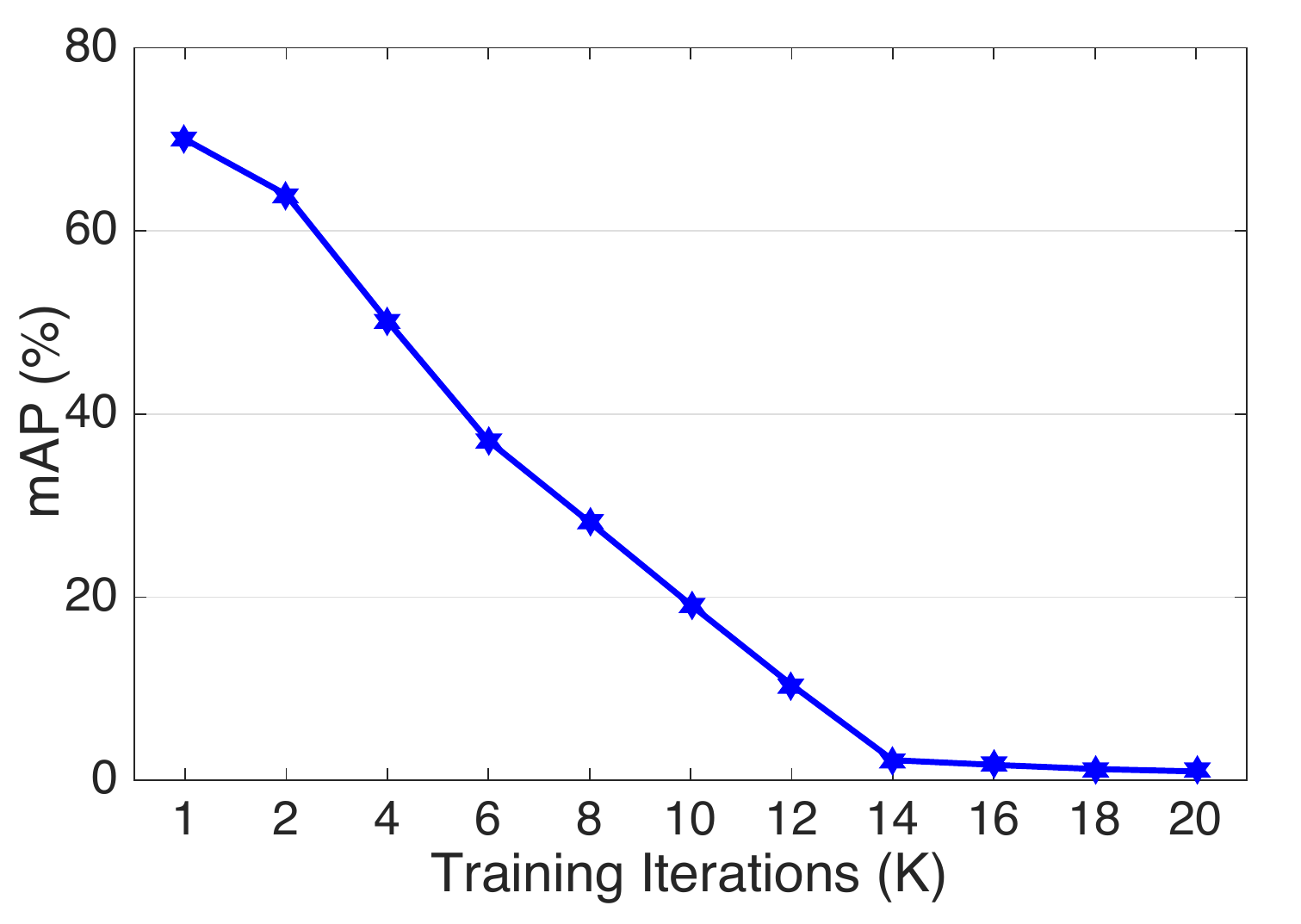}
\vspace{-10pt}
\caption{As training iterations accumulate, the falling speed of mAP gradually slow down, meaning the attack effects of \DPatch\ will saturate at a point. For $tv$, the saturate point is approximately 200k training iterations.}
\label{fig_training_iteration}
\vspace{-10pt}
\end{figure}

Table.\ref{table:faster_rcnn} demonstrates that after approximately 200k training iterations, this \DPatch\ could fool almost all the 20 classes in Pascal VOC 2007. The mAP falls down from 75.10\% to 0.98\%. We notice that at the start of training period (when training iteration is less than 40k), the falling speed of mAP is largest (see Fig.\ref{fig_training_iteration}). As the \DPatch\ accepts deeper training, its attack effect will gradually be weakened. Therefore, we can conclude that there exists a saturated point for training \DPatch. After that saturated point, increasing training iterations will no longer improve the attack effects. For $tv$, the saturate point of training iterations is about 180k.

\begin{figure}[htb]
  \centering 
  \subfigure[targeted \DPatch\ attacking Faster R-CNN]{ 
    \label{fasterrcnn_targeted} 
    \includegraphics[width=1.5in]{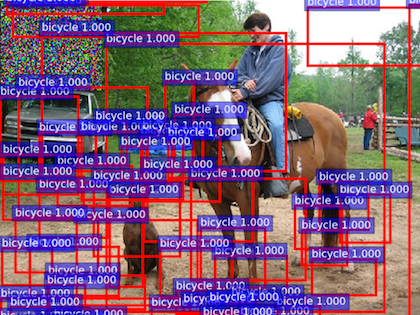}} 
  \hspace{0.05in} 
  \subfigure[targeted \DPatch\ attacking YOLO]{ 
    \label{yolo_targeted} 
    \includegraphics[width=1.5in]{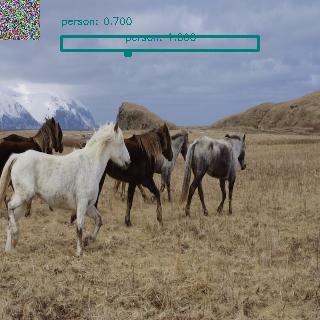}}
\vspace{-10pt}
  \caption{Successful \DPatch\ attack despite of inaccurate bounding boxes: The left is a $bike$ targeted \DPatch\ attacking Faster R-CNN, the right is a $person$ targeted \DPatch\ attacking YOLO. These wired bounding boxes are due to the inconvergence of the bounding box loss. Since our purpose is to disable the object detector but not to locate the \DPatch\ accurately, this \DPatch\ attack should be considered successful. } 
  \label{targeted_attack} 
\end{figure}

Fig.\ref{fasterrcnn_targeted} shows another \DPatch\ attacking Faster R-CNN whose targeted class is $bike$. When we apply this \DPatch\ onto an image, all the bounding boxes determined by the region proposals are disturbed. We observe that during the 200k training iterations of this $bike$ \DPatch, it takes about 100k training iterations for the class loss to decrease to approximately 0, while the bounding box loss keeps comparatively higher after 200k training iterations. That is, the detector cannot locate this \DPatch\ accurately, while its classification result is the targeted label of \DPatch. Since our purpose is to disable the object detector, considering the detector(Faster R-CNN here) has already been attacked to give wrong classification result and also the bounding box of normal objects, this targeted \DPatch\ attack should be considered successful. It is unnecessary to continue training this \DPatch\ to make the detector locate it accurately.

Fig.\ref{yolo_targeted} shows a \DPatch\ attacking YOLO whose targeted class is $person$. After we stick it to the image, the detector(YOLO here) can only give the classification as person, while the horses in the image cannot be detected.

\noindent\textbf{Fixed-sized and Randomly-located \DPatch.}
After experimenting with the fixed-located \DPatch, our next trial is to randomly shift the \DPatch\ in a scene, which means the same \DPatch\ could appear in any location of the original image. The main purpose of such practice is to evaluate the attacking efficiency of different locations where the patch is placed. If the attacking efficiency, such as spending same training iterations to get similar mAP, does not differ from each other, it is unnecessary to design a specific attack region, which means attackers can place the \DPatch\ in any area. We set the targeted class as $bike$, and kept the patch size as 20-by-20, and the attacked detector is Faster R-CNN.

Table.\ref{table:3} demonstrates the decreased mAP when Faster R-CNN is attacked by randomly-located and fixed-located \DPatch. It is notable that randomly-located one does not improve the attack result. That is, recognition accuracy(mAP) of all classes have declined to a similar value no matter where the patch is located. This result makes sense because Faster R-CNN detector will first extract thousands of region proposals all over the image, instead of finding a specific area. In this case, if the  detector search the whole image in the first step, the region with \DPatch\ will be treated as the only valid RoI. After this detection process, all the detected objects will be misclassified as the targeted class, which also has no relation with the location of \DPatch. Similarly, YOLO also firstly sees the entire image, the location of the \DPatch\ will not influence the attack results.  Therefore, we could place the \DPatch\ in an image without intentionally designing its location, which intensifies the feasibility of the attack.

\begin{table*}[htb!]
\caption{Results on Pascal VOC 2007 test set of fixed-located \DPatch\ and randomly-located \DPatch, whose targeted class is $bike$ that attacks faster R-CNN. After 200k training iterations, mAP of Pascal VOC 2007 both decreased from 75.10\% to around 25\%. Therefore, the location of \DPatch\ does not influence the attack results.}
\small
\centering
\vspace{0.1cm}
\begin{adjustbox}{width=1\textwidth}
\begin{tabular}{p{2.25cm}p{1.2cm}p{1.2cm}p{1.2cm}p{1.2cm}p{1.2cm}p{1.2cm}p{1.2cm}p{1.2cm}p{1.2cm}p{1.2cm}p{1.2cm}} 
\hline\hline
	&plane 	&bike 	&bird 	&boat 	&bottle	&bus	&car 	&cat	&chair 	&cow 	&table
\\ 
fixed-located	&25.66	&0.03	&32.48	&11.03	&8.68	&35.36	&30.19	&38.97	&13.79	&30.55	&21.7
\\ 
randomly-located	&25.69	&0.03	&32.16	&11.54	&9.68	&36.01	&30.12	&38.66	&14.09	&30.13	&22.00 
\\ \hline
&dog 	&horse 	&motor 	&person 	&plant 	&sheep 	&sofa	&train 	&tv 	&\multicolumn{2}{l}{mAP}
\\ 
    &35.77	&29.13	&19.44	&23.02	&9.33	&33.61	&36.28	&41.09	&32.16	&\textbf{25.14}
\\ 
    &36.01	&29.01	&19.39	&23.17	&9.33	&33.58	&36.30	&41.11	&32.19	&\textbf{25.51}
\\ \hline 

\end{tabular}
\end{adjustbox}
\setlength{\belowcaptionskip}{5pt}
\label{table:3}
\vspace{-0.3cm}
\end{table*}

\noindent\textbf{Multiple-Sized \DPatch.}
Since all previous targeted attacks set the \DPatch\ size to be 20-by-20, we would prefer to observe the impacts of \DPatch\ size on detectors. We add two more sizes in this test: 40-by-40 and 80-by-80. It is expected that larger-size patch can decrease mAP to a lower value. Table.\ref{table:4} validates such expectation. 

\begin{table*}[htb!]
\caption{Results on Pascal VOC 2007 test set after attacked by 20-by-20-sized, 40-by-40-sized and 80-by-80-sized \DPatch\\. The targeted class are all $cow$ that attacks Faster R-CNN. After same training iterations (200k in this case), larger sized adversarial patch can decrease mAP to a lower value.}
\small
\centering
\vspace{0.1cm}
\begin{adjustbox}{width=1\textwidth}
\begin{tabular}{p{1.75cm}p{1.2cm}p{1.2cm}p{1.2cm}p{1.2cm}p{1.2cm}p{1.2cm}p{1.2cm}p{1.2cm}p{1.2cm}p{1.2cm}p{1.2cm}} 
\hline \hline
    &plane	&bike	&bird	&boat	&bottle	&bus	&car	&cat	&chair	&cow	&table    
\\ 
20-by-20	&0.1	&0	&0	&0	&0.17	&0	&1.5	&0	&0	&4.57	&0
\\ 
40-by-40	&0	&0	&0.01	&0	&0.16	&0	&1.14	&0	&0	&4.55	&0
\\ 
80-by-80	&0	&0	&0	&0	&0.09	&0	&1	&0	&0	&3.92	&0
\\ \hline
    &dog	&horse	&motor	&person	&plant	&sheep	&sofa	&train	&tv	&\multicolumn{2}{l}{mAP}
\\ 
    &0.04	&0	&0.54	&0.21	&0.4	&0.05	&0.06	&0.03	&0	&\textbf{0.38}
\\ 
    &0.02	&0	&0.51	&0.12	&0.61	&0.02	&0	&0	&0	&\textbf{0.36}
\\ 
    &0	&0	&0.43	&0.09	&0.33	&0	&0	&0	&0	&\textbf{0.29}
\\ \hline

\end{tabular}
\end{adjustbox}
\setlength{\belowcaptionskip}{5pt}
\label{table:4}
\vspace{-0.3cm}
\end{table*}

In order to avoid the influence of training iterations, we train these three \DPatch\ es for 200k iterations and make them approach saturated points. We observe that the smallest size for valid attack differ from individual classes. For example, 20-by-20 sized \DPatch\ is robust enough to attack $bike$, $bird$, $boat$ and so on, while 80-by-80 sized one still cannot thoroughly misclassify $bottle$, $motor$, $person$ and $plant$. Therefore, we can set the \DPatch\ size according to the classes we mainly want to attack.

\begin{figure}[t]
\centering
\includegraphics[width=0.95\columnwidth]{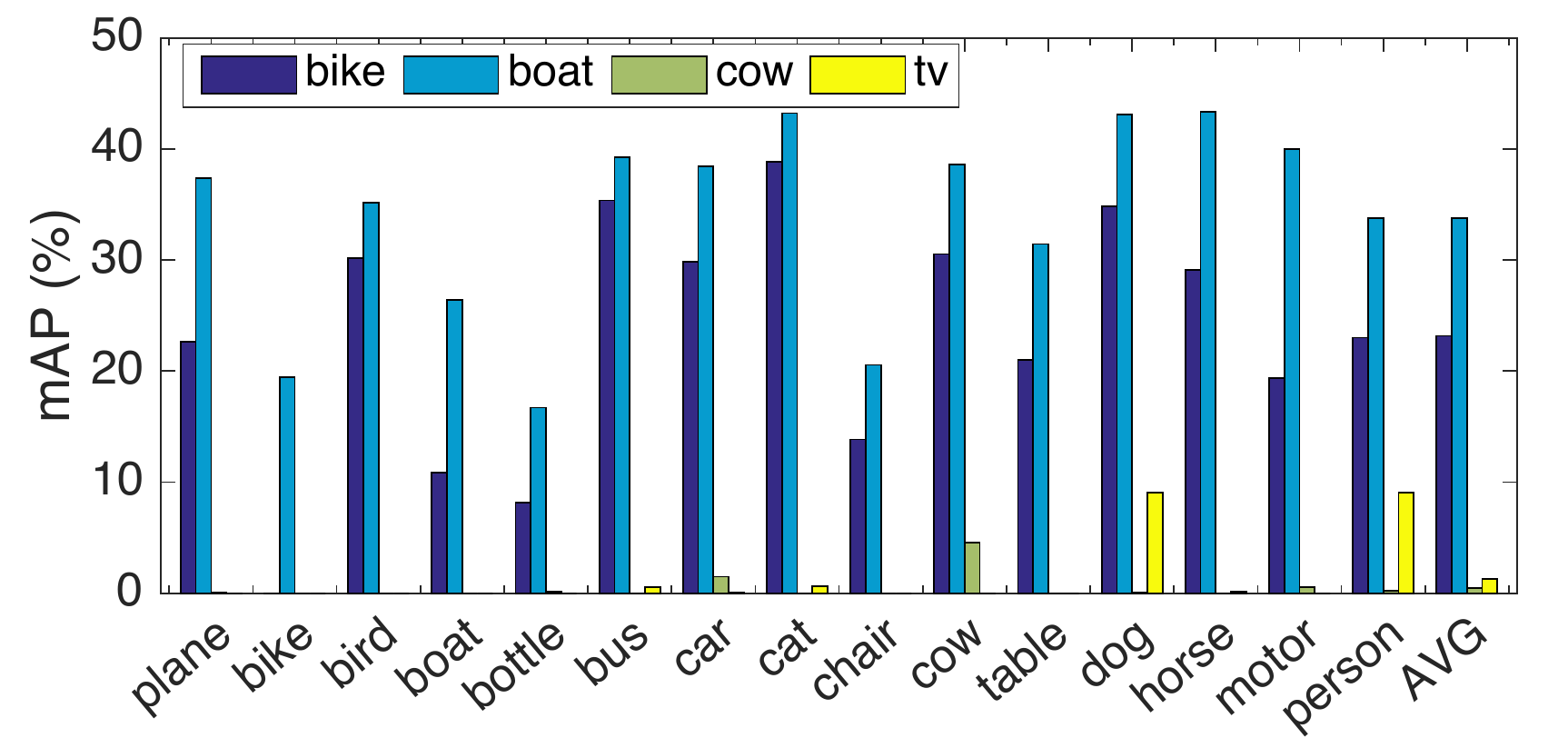}
\vspace{-10pt}
\caption{Per class mAP after attacked by \DPatch\ es of different targeted classes. $bike$ and $boat$ shrink mAP to 24.72\% and 33.50\%, while $cow$ and $tv$ shrink mAP to 0.38\% and 0.98\%. Therefore, the attack results differ from targeted classes.}
\label{fig_classes}
\vspace{-15pt}
\end{figure}

\subsection{Further Analysis}

\noindent\textbf{Training with Different Targeted Classes.}
It has been found that the bounding box loss is easier to converge for some class, e.g. $tv$, however, harder to converge for some other targeted classes like $bike$ and $person$. Therefore, we would like to explore whether \DPatch\ of different targeted classes can cause mAP drop down to a similar value after same training iterations.
  
We casually select two more classes to attack the same detector, Faster R-CNN: $boat$, $cow$ to compare with $tv$ and $bike$. Fig.\ref{fig_classes} shows that after 200k training iterations, these targeted classes cause mAP fall down to different levels. $cow$ and $tv$ both decline mAP to almost 0, while $bike$ and $boat$ shrink mAP to 24.72\% and 33.50\%. 

Based on this finding, we can conclude that $tv$ and $cow$ are more efficient to attack Faster R-CNN networks. It could be better to set the targeted class as $tv$ or $cow$ rather than $boat$ or $bike$. Hence, we can select the most efficient targeted class to train the \DPatch\ if the dataset is known.

\begin{figure}[t]
\centering
\includegraphics[width=0.7\columnwidth]{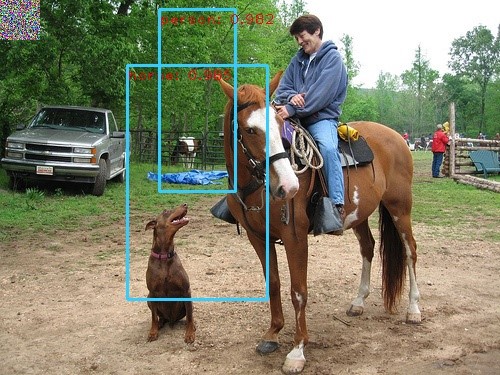}
\vspace{-10pt}
\caption{YOLO-trained \DPatch\ successfully disables Faster R-CNN.}
\label{fig_transfer}
\vspace{-10pt}
\end{figure}

\noindent\textbf{Detector Transferability of \DPatch.}
In the previous experiments, we attack the same detector as the \DPatch\ is trained on. However, it is more effective if the \DPatch\ is transferable among different detectors, specifically, we train the \DPatch\ via YOLO and it can also successfully attack Faster R-CNN. Fig.\ref{fig_transfer} shows that a \DPatch\ trained by YOLO, whose targeted class is $person$, can fool Faster R-CNN. We apply this YOLO-trained \DPatch\ to each image in the testing dataset of Pascal VOC 2007 and make Faster R-CNN work as the detector, Table.\ref{table:faster_rcnn} demonstrates that mAP of some classes, e.g. $bike$, $bird$, $bottle$, are even declined to a lower value than attack on YOLO itself, while the attack on some other classes are less effective, e.g., $plane$, $boat$.  However, mAP of all classes have considerable decrease. Similarly, Faster R-CNN trained \DPatch\ is also able to attack YOLO, see Table.\ref{table:yolo}. In such case, we can say that our \DPatch\ can be designed as a universal and black-box adversarial attack for object detectors. 

\noindent\textbf{Dataset Transferability of \DPatch.}
To investigate the transferability of \DPatch\ among different datasets, we train an untargeted DPatch using COCO  and test it on Pascal VOC trained detector. It is obvious that mAP decreases a lot after attaching the DPatch, shown in Table.\ref{coco}. Although the attack performance is not as good as the DPatch trained by the same dataset, \DPatch\ can still considerably influence the overall performance. 
\begin{table}[htb]
\centering 
\vspace{-0pt}
\caption{Transferability of \DPatch\ among different detectors and datasets. The \DPatch\ trained with COCO is able to attack the Faster R-CNN trained by Pascal VOC.}
\small
\label{coco}
\vspace{0.1cm}
\begin{tabular}{ccc}
\hline
\hline 
             & Faster R-CNN+VOC & YOLO+VOC
\\ \hline
no DPatch    & 75.10            & 65.70
\\
COCO-trained DPatch & 28.00 & 24.34
\\
VOC-trained DPatch  & 2.90 & 0.00
\\ \hline
\end{tabular}
\vspace{-0pt}
\end{table}

\begin{figure} 
  \centering 
  \subfigure[YOLO-extracted RoIs with \DPatch]{ 
    \label{fig_roi_yolo_a} 
    \includegraphics[width=1.5in]{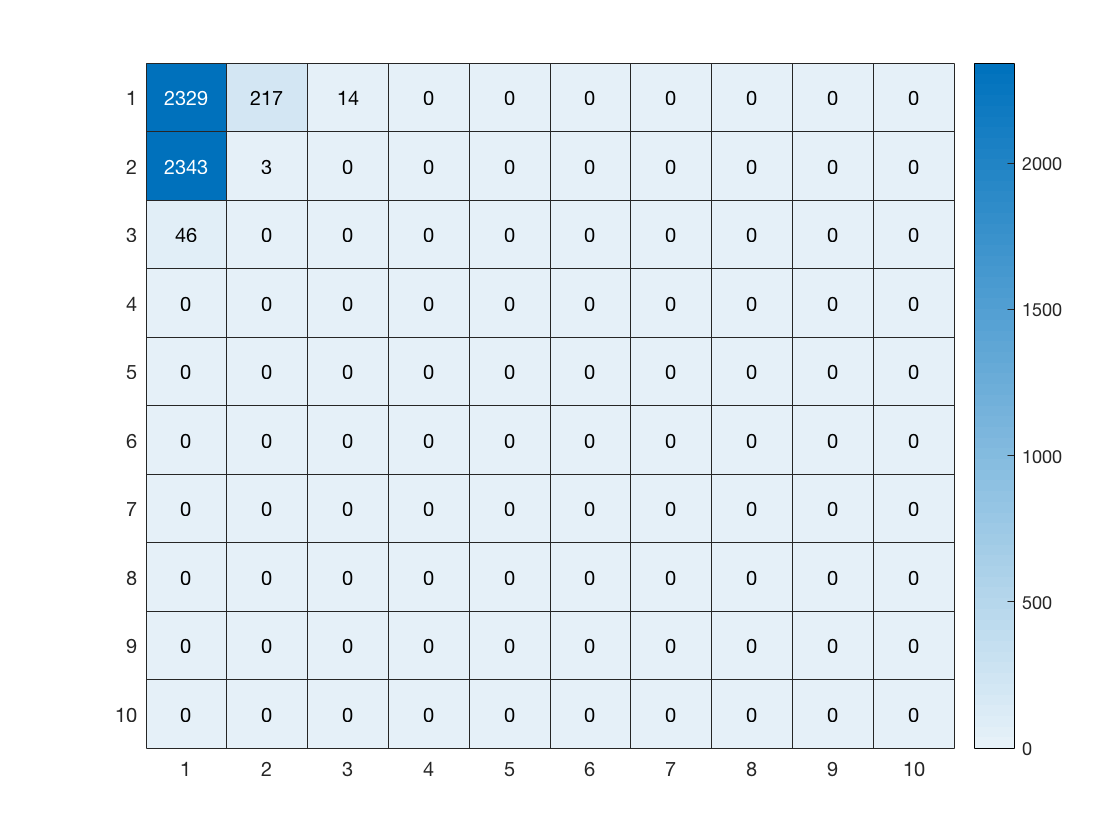}} 
  \hspace{0.05in} 
  \subfigure[YOLO-extracted RoIs without \DPatch]{ 
    \label{fig_roi_yolo_b} 
    \includegraphics[width=1.5in]{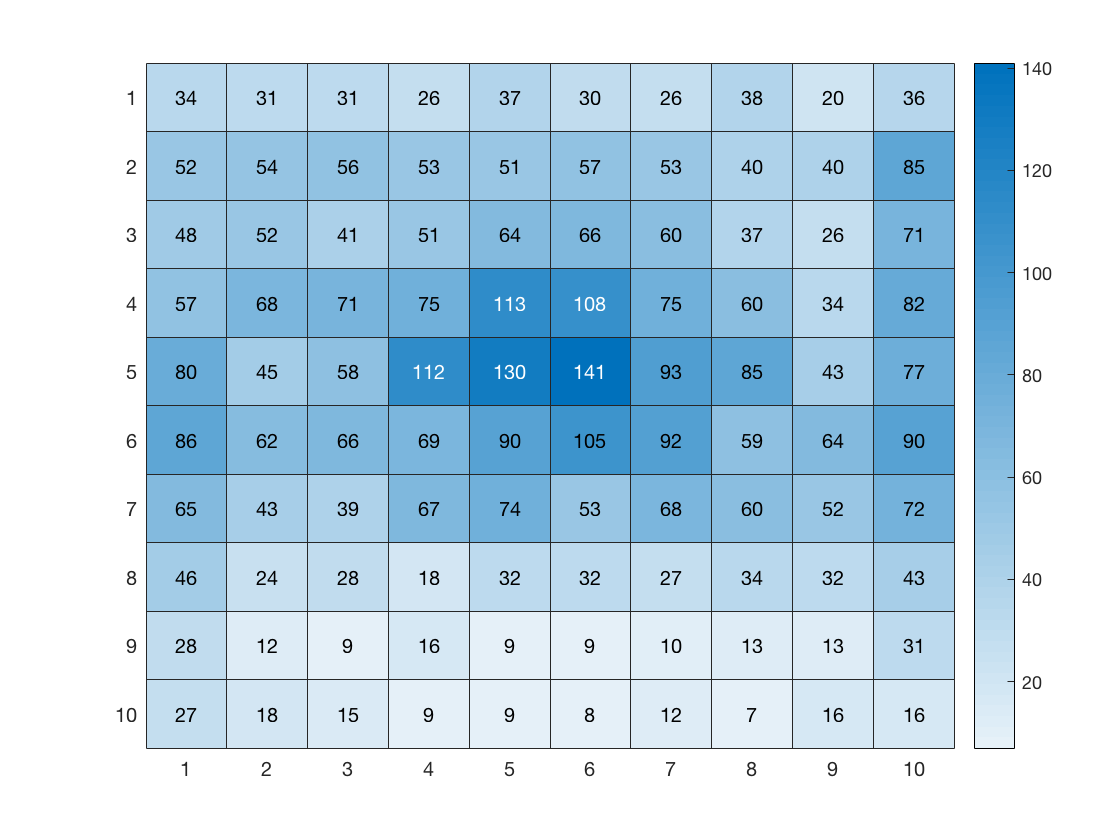}} 
  \caption{The left is YOLO-extracted RoIs after adding \DPatch\ in the left upper corner of each image. The number in each grid is the frequency of the RoI appears in it. Obviously, RoIs gather in the area where \DPatch\ locates. On contrary, the right is extracted RoIs without \DPatch. RoIs are distributed over all of the grids. } 
  \label{fig_roi_yolo} 
\end{figure}

\noindent\textbf{Why \DPatch\ Works.}
As explained before, our main purpose is to train a \DPatch\ pattern that once attached to the input image, the RoIs extracted by the detector is the region that \DPatch\ occupies. After extracting RoIs, detector performs classification tasks and bounding box regressions on these RoI areas while ignore others. In such case, if the \DPatch\ attacks the detector successfully, most extracted RoIs should gather in the region where we attach \DPatch. To verify this inference, we sum up the RoIs to find the most frequent area that RoI appears. 
As shown in Fig.\ref{fig_roi_yolo}, the number in each grid represents the frequency that RoI occupies it. We attach a \DPatch\ in the left upper corner of each image in the testing dataset, and let YOLO work as the detector. Fig.\ref{fig_roi_yolo_a} verifies that YOLO-extracted RoIs are concentrated on grid 1 and grid 2, where our \DPatch\ locates. On contrary, Fig.\ref{fig_roi_yolo_b} shows that without \DPatch, YOLO-extracted RoIs are distributed over all grids.

\begin{figure} 
  \centering 
  \subfigure[Faster R-CNN extracted RoIs with \DPatch]{ 
    \label{fig_roi_faster_a} 
    \includegraphics[width=1.25in]{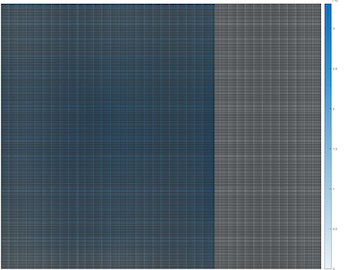}} 
  \hspace{0.05in} 
  \subfigure[Faster R-CNN extracted RoIs without \DPatch]{ 
    \label{fig_roi_faster_b} 
    \includegraphics[width=1.2in]{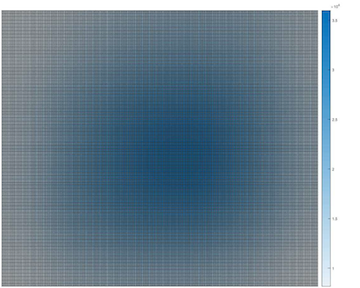}} 
  \caption{Darker the color, higher the frequency of RoI being predicted here. The left is Faster R-CNN extracted RoIs after adding \DPatch, obviously concentrated on the left, while the right is Faster R-CNN extracted RoIs without \DPatch, more evenly distributed. } 
  \label{fig_roi_faster} 
\end{figure}

Similarly, Fig.\ref{fig_roi_faster} demonstrates the distribution of RoIs extracted by Faster R-CNN with and without \DPatch. It is clear that the regions that \DPatch\ occupies are considered as RoIs more frequently, as shown in Fig.\ref{fig_roi_faster_a}. Instead, RoIs distribute much more evenly without the influence of \DPatch, shown in Fig.\ref{fig_roi_faster_b}.

Such analysis verifies that our \DPatch\ performs a successful attack towards object detectors, mainly because all the RoIs is actually occupied by the \DPatch\ , instead of normal objects.

\section{Conclusions}
In this work, we successfully attack modern object detectors using our proposed \DPatch, where the adversarial attack is performed by learning and embedding a small patch in the input image.
Compared to prior works, our \DPatch\ has several appealing properties:
(1) \DPatch\ can perform effective attacks against mainstream modern detector architectures, such as the two-stage detector Faster R-CNN and the one-stage detector YOLO;
(2) \DPatch\ is small in size and its attacking effect is location-independent, making it very practical to implement real-world attacks;
(3) \DPatch\ demonstrates great transferability between different detector architectures as well as training datasets. For example, \DPatch\ that is trained on Faster R-CNN can effectively attack YOLO, and vice versa. 
Our experiments imply that \DPatch\ can perform effective attacks under black-box setup, i.e., even without the knowledge of the attacked network's architectures and parameters.
The successful realization of \DPatch\ also illustrates the intrinsic vulnerability of the modern detector architectures to such patch-based adversarial attacks.

\bibliographystyle{aaai}
\bibliography{reference.bib}

\end{document}